\documentclass[superscriptaddress,
prx,
preprintnumbers,
amsmath,
amssymb,
altaffilletter,
floatfix,
]{revtex4-2}
\usepackage{graphicx}% Include figure files
\usepackage{dcolumn}% Align table columns on decimal point
\usepackage{bm}% bold math
\usepackage{caption}
\usepackage{subcaption}
\usepackage{url}
\usepackage{orcidlink}
\def\mjd{\textsc{Majorana Demonstrator}}

\def\ge{$^{76}$Ge}

\def\vbb{$0\nu\beta\beta$}

\def\th{$^{228}$Th}

\begin{document}

\preprint{APS/123-QED}

\title{{\sc Majorana Demonstrator} Data Release for AI/ML Applications}% Force line breaks with \\

% \author{Ann Author}
%  \altaffiliation[Also at ]{Physics Department, XYZ University.}%Lines break automatically or can be forced with \\
%  \email{Second.Author@institution.edu}
% \affiliation{%
%  Authors' institution and/or address\\
%  This line break forced with \textbackslash\textbackslash
% }%

% \collaboration{MUSO Collaboration}%\noaffiliation

%  \homepage{http://www.Second.institution.edu/~Charlie.Author}
% \affiliation{
%  Second institution and/or address\\
%  This line break forced% with \\
% }%
% \affiliation{
%  Third institution, the second for Charlie Author
% \affiliation{%
%  Authors' institution and/or address\\
%  This line break forced with \textbackslash\textbackslash
% }%
\newcommand{\ITEP}{National Research Center ``Kurchatov Institute'', Kurchatov Complex of Theoretical and Experimental Physics, Moscow, 117218 Russia}
\newcommand{\JINR}{Joint Institute for Nuclear Research, Dubna, 141980 Russia} 
\newcommand{\lbnl}{Nuclear Science Division, Lawrence Berkeley National Laboratory, Berkeley, CA 94720, USA}
\newcommand{\lbnle}{Engineering Division, Lawrence Berkeley National Laboratory, Berkeley, CA 94720, USA}
\newcommand{\lanl}{Los Alamos National Laboratory, Los Alamos, NM 87545, USA}
\newcommand{\queens}{Department of Physics, Engineering Physics and Astronomy, Queen's University, Kingston, ON K7L 3N6, Canada}
\newcommand{\uw}{Center for Experimental Nuclear Physics and Astrophysics, and Department of Physics, University of Washington, Seattle, WA 98195, USA}
\newcommand{\unc}{Department of Physics and Astronomy, University of North Carolina, Chapel Hill, NC 27514, USA}
\newcommand{\duke}{Department of Physics, Duke University, Durham, NC 27708, USA}
\newcommand{\ncsu}{Department of Physics, North Carolina State University, Raleigh, NC 27695, USA}	
\newcommand{\ornl}{Oak Ridge National Laboratory, Oak Ridge, TN 37830, USA}
\newcommand{\ou}{Research Center for Nuclear Physics, Osaka University, Ibaraki, Osaka 567-0047, Japan}
\newcommand{\pnnl}{Pacific Northwest National Laboratory, Richland, WA 99354, USA}
\newcommand{\ttu}{Tennessee Tech University, Cookeville, TN 38505, USA}
\newcommand{\sdsmt}{South Dakota Mines, Rapid City, SD 57701, USA}
\newcommand{\usc}{Department of Physics and Astronomy, University of South Carolina, Columbia, SC 29208, USA}
\newcommand{\usd}{Department of Physics, University of South Dakota, Vermillion, SD 57069, USA}  
\newcommand{\ut}{Department of Physics and Astronomy, University of Tennessee, Knoxville, TN 37916, USA}
\newcommand{\tunl}{Triangle Universities Nuclear Laboratory, Durham, NC 27708, USA}
\newcommand{\mpi}{Max-Planck-Institut f\"{u}r Physik, M\"{u}nchen, 80805, Germany}
\newcommand{\tum}{Physik Department and Excellence Cluster Universe, Technische Universit\"{a}t, M\"{u}nchen, 85748 Germany}
\newcommand{\williams}{Physics Department, Williams College, Williamstown, MA 01267, USA}
\newcommand{\ciemat}{Centro de Investigaciones Energ\'{e}ticas, Medioambientales y Tecnol\'{o}gicas, CIEMAT 28040, Madrid, Spain}
\newcommand{\iu}{IU Center for Exploration of Energy and Matter, and Department of Physics, Indiana University, Bloomington, IN 47405, USA}

\author{I.J.~Arnquist}\affiliation{\pnnl} 
\author{F.T.~Avignone~III}\affiliation{\usc}\affiliation{\ornl}
\author{A.S.~Barabash\,\orcidlink{0000-0002-5130-0922}}\affiliation{\ITEP}
\author{C.J.~Barton}\affiliation{\usd}	
\author{K.H.~Bhimani}\affiliation{\unc}\affiliation{\tunl} 
\author{E.~Blalock}\affiliation{\ncsu}\affiliation{\tunl} 
\author{B.~Bos}\affiliation{\unc}\affiliation{\tunl} 
\author{M.~Busch}\affiliation{\duke}\affiliation{\tunl}	
\author{M.~Buuck}\altaffiliation{Present address: SLAC National Accelerator Laboratory, Menlo Park, CA 94025, USA}\affiliation{\uw} 
\author{T.S.~Caldwell}\affiliation{\unc}\affiliation{\tunl}	
\author{Y.-D.~Chan}\affiliation{\lbnl}
\author{C.D.~Christofferson}\affiliation{\sdsmt} 
\author{P.-H.~Chu\,\orcidlink{0000-0003-1372-2910}}\affiliation{\lanl} 
\author{M.L.~Clark}\affiliation{\unc}\affiliation{\tunl} 
\author{C.~Cuesta\,\orcidlink{0000-0003-1190-7233}}\affiliation{\ciemat}	
\author{J.A.~Detwiler\,\orcidlink{0000-0002-9050-4610}}\affiliation{\uw}	
\author{Yu.~Efremenko}\affiliation{\ut}\affiliation{\ornl}
\author{H.~Ejiri}\affiliation{\ou}
\author{S.R.~Elliott\,\orcidlink{0000-0001-9361-9870}}\affiliation{\lanl}
\author{N.~Fuad\,\orcidlink{0000-0002-5445-2534}}\affiliation{\iu}
\author{G.K.~Giovanetti}\affiliation{\williams}  
\author{M.P.~Green\,\orcidlink{0000-0002-1958-8030}}\affiliation{\ncsu}\affiliation{\tunl}\affiliation{\ornl}   
\author{J.~Gruszko\,\orcidlink{0000-0002-3777-2237}}\affiliation{\unc}\affiliation{\tunl} 
\author{I.S.~Guinn\,\orcidlink{0000-0002-2424-3272}}\affiliation{\unc}\affiliation{\tunl} 
\author{V.E.~Guiseppe\,\orcidlink{0000-0002-0078-7101}}\affiliation{\ornl}	
\author{C.R.~Haufe}\affiliation{\unc}\affiliation{\tunl}	
\author{R.~Henning}\affiliation{\unc}\affiliation{\tunl}
\author{D.~Hervas~Aguilar}\affiliation{\unc}\affiliation{\tunl} 
\author{E.W.~Hoppe}\affiliation{\pnnl}
\author{A.~Hostiuc}\affiliation{\uw} 
\author{M.F.~Kidd}\affiliation{\ttu}	
\author{I.~Kim}\affiliation{\lanl} \altaffiliation{Present address: Lawrence Livermore National Laboratory, Livermore, CA 94550, USA} 
\author{R.T.~Kouzes}\affiliation{\pnnl}
\author{T.E.~Lannen~V}\affiliation{\usc} 
\author{A.~Li\,\orcidlink{0000-0002-4844-9339}}\affiliation{\unc}\affiliation{\tunl} 
\author{J.M. L\'opez-Casta\~no}\affiliation{\ornl} 
\author{R.D.~Martin}\affiliation{\queens}	
\author{R.~Massarczyk}\affiliation{\lanl}		
\author{S.J.~Meijer\,\orcidlink{0000-0002-1366-0361}}\affiliation{\lanl}	
\author{S.~Mertens}\affiliation{\mpi}\affiliation{\tum}	
\author{T.K.~Oli}\altaffiliation{Present address: Argonne National Laboratory, Lemont, IL 60439, USA}\affiliation{\usd}  
\author{L.S.~Paudel\,\orcidlink{0000-0003-3100-4074}}\affiliation{\usd} 
\author{W.~Pettus\,\orcidlink{0000-0003-4947-7400}}\affiliation{\iu}	
\author{A.W.P.~Poon\,\orcidlink{0000-0003-2684-6402}}\affiliation{\lbnl}
\author{B.~Quenallata}\affiliation{\ciemat} 
\author{D.C.~Radford}\affiliation{\ornl}
\author{A.L.~Reine\,\orcidlink{0000-0002-5900-8299}}\affiliation{\unc}\affiliation{\tunl}	
\author{K.~Rielage\,\orcidlink{0000-0002-7392-7152}}\affiliation{\lanl}
\author{N.W.~Ruof\,\orcidlink{0000-0002-0477-7488}}\affiliation{\uw}\altaffiliation{Present address: Lawrence Livermore National Laboratory, Livermore, California 94550, USA} 
\author{D.C.~Schaper}\affiliation{\lanl}
\author{S.J.~Schleich\,\orcidlink{0000-0003-1878-9102}}\affiliation{\iu}
\author{D.~Tedeschi}\affiliation{\usc}		
\author{R.L.~Varner\,\orcidlink{0000-0002-0477-7488}}\affiliation{\ornl}  
\author{S.~Vasilyev}\affiliation{\JINR}	
\author{S.L.~Watkins}\affiliation{\lanl} 
\author{J.F.~Wilkerson\,\orcidlink{0000-0002-0342-0217}}\affiliation{\unc}\affiliation{\tunl}\affiliation{\ornl}    
\author{C.~Wiseman\,\orcidlink{0000-0002-4232-1326}}\affiliation{\uw}		
\author{W.~Xu}\affiliation{\usd} 
\author{C.-H.~Yu\,\orcidlink{0000-0002-9849-842X}}\affiliation{\ornl}
\author{B.X.~Zhu}\altaffiliation{Present address: Jet Propulsion Laboratory, California Institute of Technology, Pasadena, CA 91109, USA}\affiliation{\lanl}

\collaboration{{\sc{Majorana}} Collaboration}
\noaffiliation
% \collaboration{\textsc{Majorana Demonstrator} Collaboration}%\noaffiliation

\date{\today}% It is always \today, today,

%\tableofcontents

\maketitle

The enclosed data release consists of a subset of the $^{228}$Th calibration data from the \textsc{Majorana Demonstrator} experiment. Each \textsc{Majorana} event is accompanied by raw Germanium detector waveforms, pulse shape discrimination cuts, and calibrated final energies, all shared in an HDF5 file format along with relevant metadata.  This release is specifically designed to support the training and testing of Artificial Intelligence and Machine Learning (AI/ML) algorithms upon our data. This document is structured as follows. Section~\ref{sec:content} provides an overview of the dataset's content and format;  Section~\ref{sec:access} outlines the location of this dataset and the method for accessing it; Section~\ref{sec:NPML} presents the NPML Machine Learning Challenge associated with this dataset; Section~\ref{sec:disclamir} contains a disclaimer from the \mjd~regarding the use of this dataset; Appendix~\ref{app:detail} contains technical details of this data release. Please direct questions about the material provided within this release to \url{liaobo77@ucsd.edu} (A. Li).

\section{Dataset description}\label{sec:content}
Artificial Intelligence and Machine Learning (AI/ML) have become increasingly integral to contemporary society, and a key component of their development lies in the training and test datasets.  Simultaneously, nuclear physics experiments generate substantial volumes of high-quality data, complemented by robust labels derived from dedicated physics analysis for each data point. As a member in the experimental nuclear physics community, the \textsc{Majorana Demonstrator} experiment produces high-quality time series data accompanied by rigorous, well-understood analysis labels. Acknowledging the demand from the AI/ML community and the encouragement from funding agencies~~\cite{nelson,NAIRR}, the \mjd~collaboration has taken the decision to release a portion of its detector waveforms with corresponding analysis labels to support the training and test of AI/ML algorithms.

\subsection{Detector waveforms}\label{subsec:waveforms}
\begin{figure*}[hbt!]
    \centering
    \includegraphics[width=0.95\linewidth,,trim={7pc 0pc 7pc 5pc},clip]{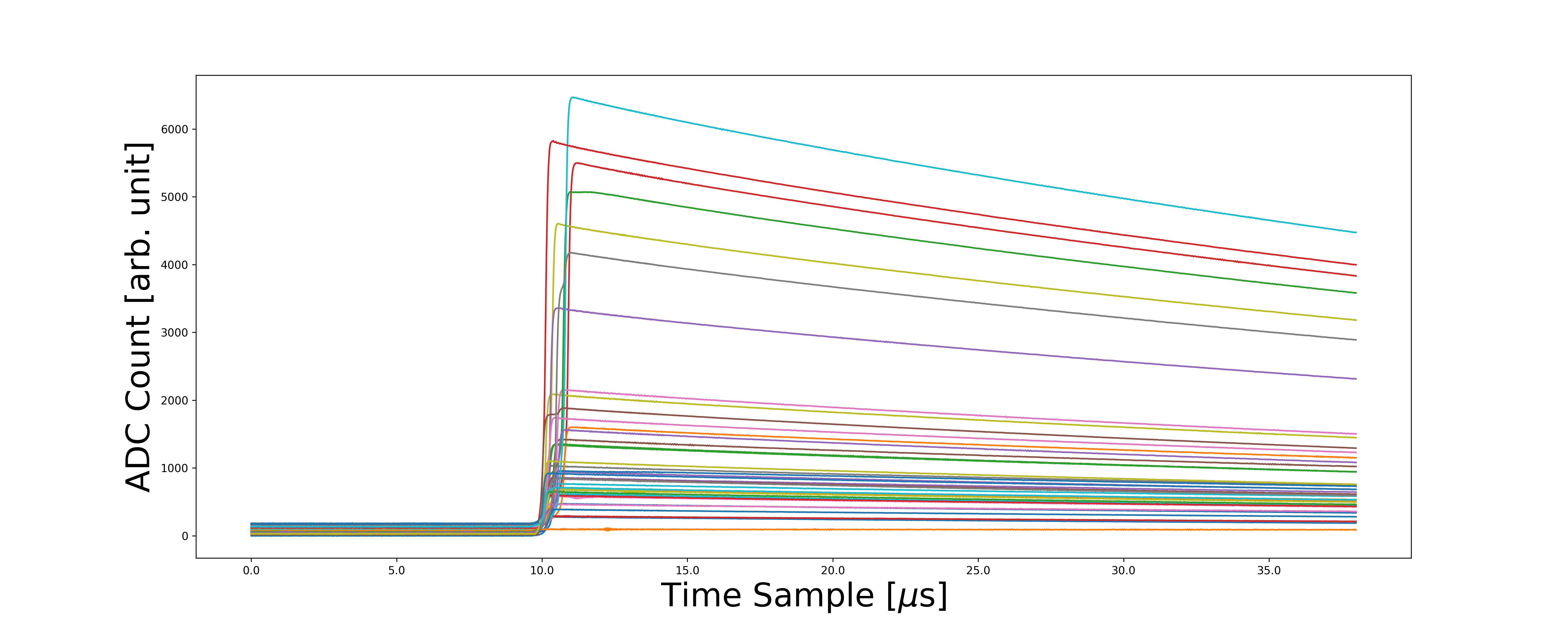}
    \caption{This plot includes samples of \mjd~waveforms from this data relase. The x-axis represents the time samples in units of nanoseconds, and the y-axis corresponds to the count of Analog-to-Digital Converter (ADC) channels. Due to the data structure of \mjd~waveforms, the rising edge is fully sampled, while the baseline region and the falling edge are pre-summed for every 4 samples, and recovered by linear interpolation.}
    \label{fig:mjdwf}
\end{figure*}

The \mjd~experiment searched for neutrinoless double-$\beta$ decay (\vbb) of \ge~using modular arrays of 56 high-purity Ge (HPGe) detectors. These detectors produce short time series data, or waveforms, as illustrated in Figure~\ref{fig:mjdwf}. These waveforms can be divided into three distinct regions. The \texttt{baseline} region precedes $\sim$10\,$\mu s$ and contains only baseline noise. The baseline's height and noise amplitude could vary among different detectors and drift among different runs. When a particle enters HPGe detectors, the ADC count suddenly increase to a higher value, producing the \texttt{rising edge} of the waveform. The relative rise in ADC count with respect to baseline is approximately proportional to the energy of the incident particle, and the shape of the rising edge provides information about the type of incident particles. Following the rising edge is the \texttt{decaying tail} region  where the waveform exhibits a long, exponentially decaying tail.

\subsection{Analysis labels}\label{subsec:analysis_labels}
To obtain field-leading physics results~\cite{mjd_prl, mjd_prl_old, mjd_prc, mjd_bosdm, mjd_lip, mjd_prc, mjd_excited, MAJORANA:2022rxr, pepv, alphan}, the \mjd~collaboration has invested significant effort in comprehending the underlying detector physics inherent within each waveform. This understanding is condensed into a set of analysis labels. Each waveform's analysis label is fine-tuned based on our current best knowledge to accurately represent the characteristics of incident particles. In this section, we provide a concise overview of these analysis labels, aiming to offer sufficient information for AI/ML users. For a more comprehensive understanding of the underlying physics, we refer the reader to the relevant publications cited within each bullet point. The analysis labels can be divided into two categories:
\begin{itemize}
    \item \texttt{energy\_label}: In Section~\ref{subsec:waveforms}, we mentioned that the relative rise in ADC count with respect to the baseline is approximately, rather than exactly, proportional to the energy of the incident particle. This is due to several detector-specific effects that could influence this proportion~\cite{charge_trapping}. To achieve the highest possible precision in reproducing particle energy, the collaboration has applied various corrections to compensate for these effects. This includes performing detector-by-detector and run-by-run energy calibration~\cite{mjd_ecal}. The resulting finely-tuned energy of each waveform is then recorded as the energy label.
    \item \texttt{psd\_label}: PSD stands for pulse shape discrimination. PSD analysis leverages information from the rising edge and the decaying tail of waveforms to distinguish desired signal-like particles from background-like particles. In the \mjd~analysis, four significant PSD labels are utilized. Similar to the energy label, each of these four PSD labels undergoes various corrections, fine-tuning on a detector-by-detector basis, and undergoes a comprehensive uncertainty evaluation. The four major PSD labels used in the \mjd~analysis are:
    \begin{enumerate}
        \item \texttt{psd\_label\_low\_avse}: remove waveforms with low AvsE value to reject multi-site backgrounds. AvsE corresponds to the Current Amplitude vs. Energy parameter~\cite{avse_paper}.
        \item \texttt{psd\_label\_high\_avse}: remove waveforms with high AvsE value to reject surface events near the point contact~\cite{avse_paper,bdt_paper}
        \item \texttt{psd\_label\_dcr}: reject surface $\alpha$ events based on the decaying tail. DCR corresponds to the Delayed Charge Recovery parameter~\cite{TUBE_paper}.
        \item \texttt{psd\_label\_lq}: reject partial charge deposition events in the transition layer based on the turning region between rising edge and decaying tail. LQ corresponds to the Late Charge parameter~\cite{mjd_prl}.
    \end{enumerate}
\end{itemize}

\subsection{Dataset content}\label{subsec:analysis_labels}
The proposed dataset contains a randomly selected 1\% subset of DS6 \th~calibration runs of \mjd~detectors. The full dataset, after data cleaning and run selection cuts, contains 3\,193\,486 data points. Each data point contains 9 fields listed in Table~\ref{tab:osfield}. Despite the detector waveform and analysis label discussed in Section~\ref{subsec:waveforms} and \ref{subsec:analysis_labels}, there are 4 additional fields that have not been discussed before:
\begin{itemize}
    \item \texttt{tp0}: index of the start of the rising edge, estimated using an asymmetric trapezoidal filter.
    \item \texttt{run\_number}: the calibration run which the waveform comes from. Note that the baseline amplitude may drift among different runs. 
    \item \texttt{detector}: specifies the detector from which each data point originates. This id is unique for each detector.
    \item \texttt{id}: An unique id for every data point in the dataset, starting from 0 up to 3\,193\,485.
\end{itemize}

\begin{center}
\renewcommand{\arraystretch}{1.7}
\setlength{\tabcolsep}{10pt}
\begin{table}[hbt]
\centering
\caption{Description of information contained in each data point of this release.} 
\begin{tabular}{cccl}%{@{}l*{30}{l}}
\hline
Field & Description&  Data Type & Note \\
\hline
\texttt{raw\_waveform} & Detector Waveform& \texttt{array(size=(3800,) dtype=float)}&\\
\hline
\texttt{energy\_label} & Analysis Label& \texttt{float}& \\
\texttt{psd\_label\_low\_avse} & Analysis Label& \texttt{binary}& 1 means accepted, 0 means rejected \\
\texttt{psd\_label\_high\_avse} & Analysis Label& \texttt{binary}& 1 means accepted, 0 means rejected \\
\texttt{psd\_label\_dcr} & Analysis Label& \texttt{binary}& 1 means accepted, 0 means rejected \\
\texttt{psd\_label\_lq} & Analysis Label& \texttt{binary}& 1 means accepted, 0 means rejected \\
\hline
\texttt{tp0} & Analysis Parameter & \texttt{integer}& Start of the rising edge\\
\hline
\texttt{detector} & Metadata& \texttt{integer}& unique ID for each detector \\
\texttt{run\_number} & Metadata& \texttt{integer}& unique ID for each run \\
\texttt{id} & Metadata& \texttt{integer}& unique ID for each data point \\
\hline
% \hline
% \texttt{psd\_label\_} & \begin{tabular}{@{}c@{}} 4 Labels:\\ \texttt{Low AvsE/High AvsE/DCR/LQ}\end{tabular} & 1 Combined Label&\begin{tabular}{@{}l@{}} $\cdot$ All labels are binary numbers (1 for accept, 0 for reject)\\ instead of the actual PSD values \\  $\cdot$ Combined label will be 1 if all 4 PSD cuts accepts \\and 0 otherwise\end{tabular}\\
% \hline
% \texttt{energy\_label} & \multicolumn{2}{|c|}{\texttt{Final\_Energy}}&The calibrated final energy of each waveform\\
% \hline
% \texttt{event\_id} & 1,2, ... 3\,193\,486 & 1,2, ... 743\,282 &Assigned unique ID for each event\\
% \hline
% \texttt{detector\_channel} & Actual Channel No. & Masked Channel No.&\begin{tabular}{@{}l@{}} Masked channel No. means we map each of detector\\ channel  to a new number, i.e. Ch. 626 $\xrightarrow[]{}$ 1, Ch. 660 \\$\xrightarrow[]{} 2$, The mapping will be kept proprietary \end{tabular}\\
% \hline
\end{tabular}
\label{tab:osfield}
\end{table}
\end{center}

\par
\begin{figure*}[hbt!]
    \centering
    \includegraphics[width=0.95\linewidth,,trim={15pc 0pc 12pc 5pc},clip]{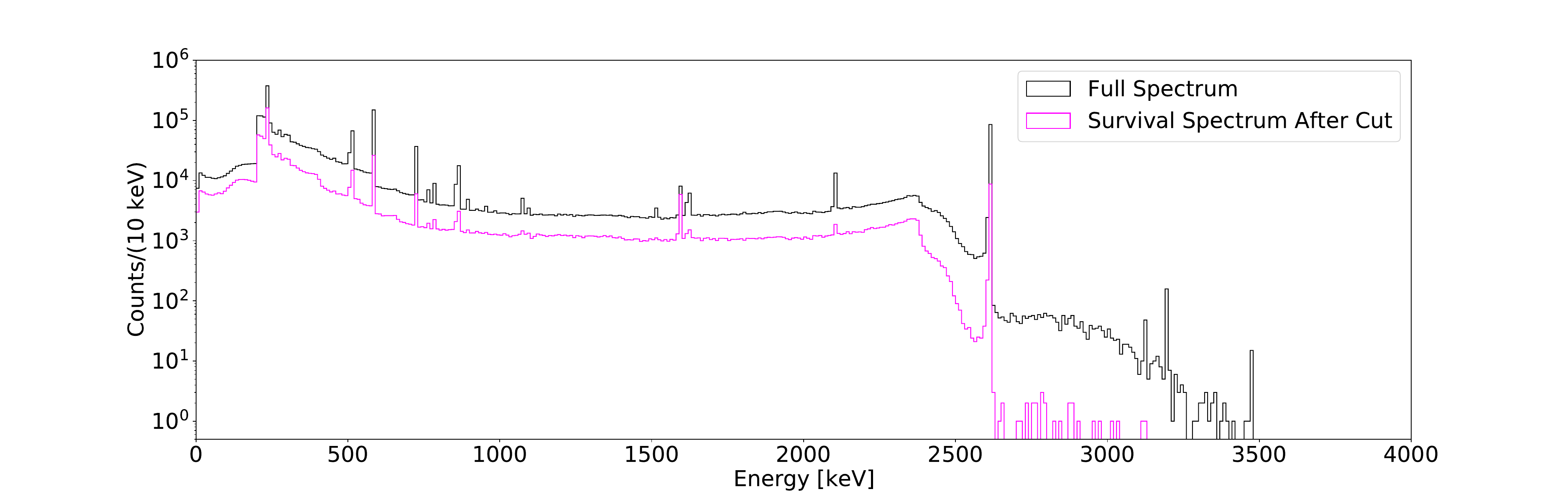}
    \caption{The energy spectrum of all events (3.2 million) and the energy spectrum of events surviving all PSD cuts (1.2 million).}
    \label{fig:spectrum}
\end{figure*}

 The energy spectrum of the full dataset and the dataset after all cuts are shown in Figure~\ref{fig:spectrum}. The sharp, step-like feature around 100\,keV is due to a cut used in \mjd~data processing. The dataset is split into 3 portions with 75\%:20\%:5\% ratio. The 75\% portion is the training dataset, and the 20\% subset is the test dataset, both contains all fields in Table~\ref{tab:osfield}. The 5\% subset is exclusively designated as the NPML challenge dataset, primarily tailored for the NPML challenge. It solely comprises the following fields: \texttt{raw\ waveform}, \texttt{detector}, \texttt{run\_number}, \texttt{tp0}, and \texttt{id}. Notably, all analysis labels are omitted from this subset to create a distinct evaluation set for the NPML challenge. Further details concerning the NPML challenge can be found in Section~\ref{sec:NPML}.

\section{Dataset access}\label{sec:access}
\par
\begin{figure}[htb!]
    \includegraphics[width=0.4\linewidth,,trim={0pc 0pc 0pc 0pc},clip]{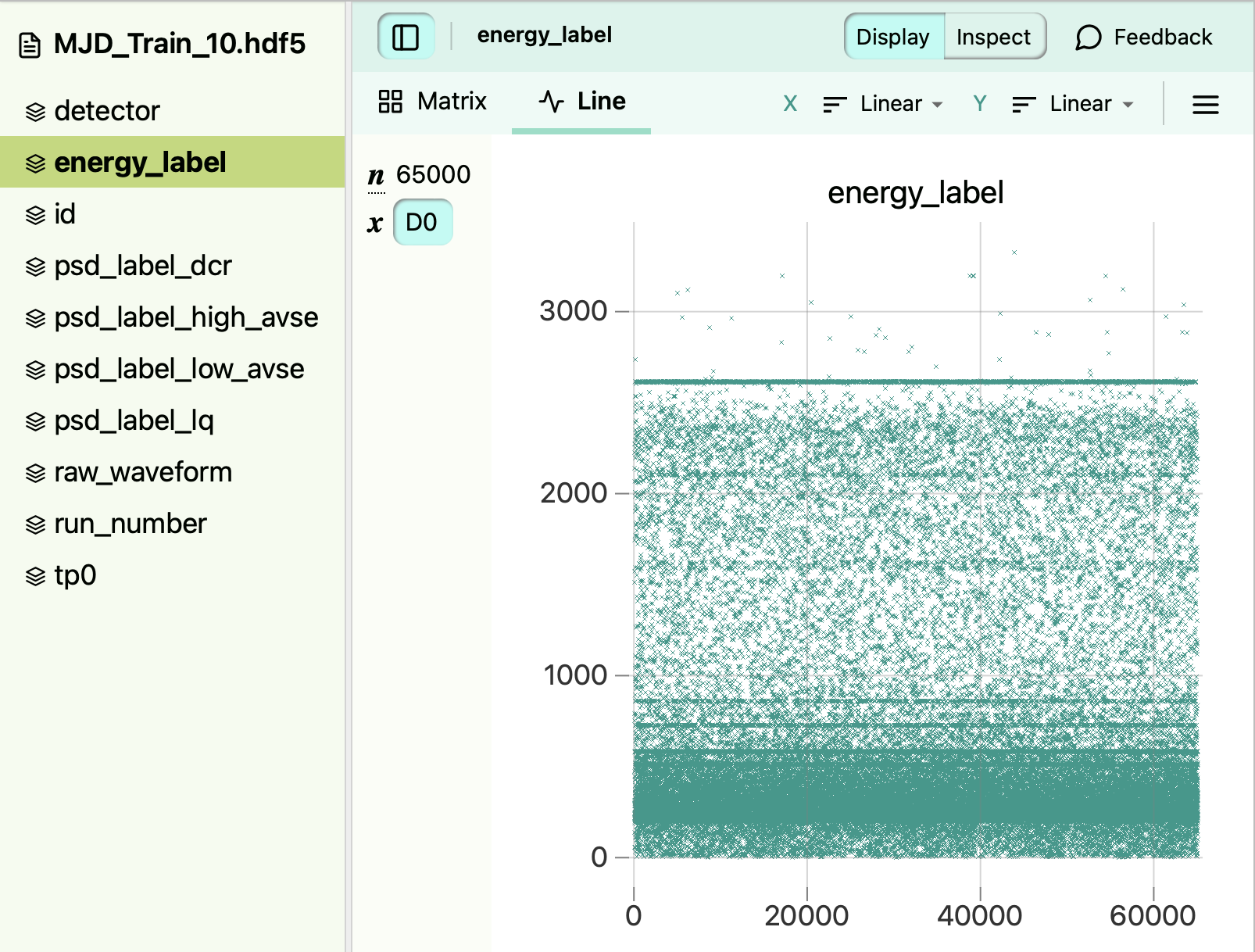}
    \hspace{2em}
    \includegraphics[width=0.4\linewidth,,trim={0pc 0pc 0pc 0pc},clip]{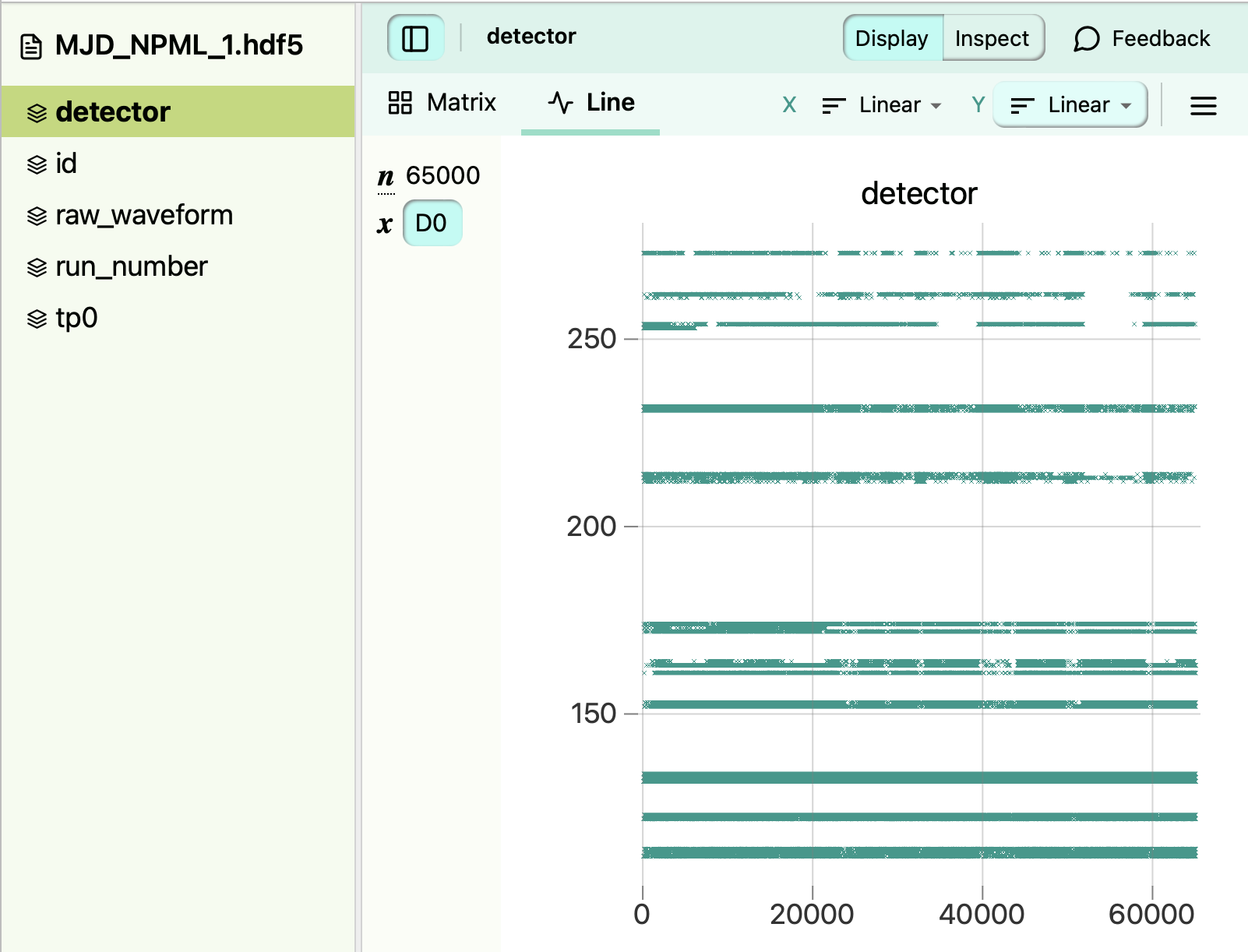}
    \caption{Left: the HDF5 data structure of the train and test dataset; Right: the HDF5 data structure of the NPML challenge dataset.}
    \label{fig:hdf5}
\end{figure}

 The dataset described above will be stored in the HDF5 format (\texttt{.hdf5}). HDF5 files are specifically designed to accommodate large and diverse datasets, and it can be easily accessed in python via the \texttt{h5py} package (\url{https://www.h5py.org}). Figure~\ref{fig:hdf5} provides a detailed overview of both the full and masked HDF5 files.

\par The train, test, and NPML dataset will contains events from all calibration runs. Each dataset is then chunked into various equal-sized subsets with chunk size of 65\,000 plus one smaller subset with all residual data points. each subset is then stored as an HDF5 file. The chunk size of 65\,000 is deliberately selected to constrain the size of a single \texttt{hdf5} file to below 2\,Gb. Table~\ref{tab:ds_summary} provides an overall summary of the dataset to be released. Users need to create an account for the DataPlanet system in order to access the data.

\begin{center}
\renewcommand{\arraystretch}{1.7}
\setlength{\tabcolsep}{5pt}
\begin{table}[hbt]
\centering
\caption{Summary of critical informations about this data release. In line 5 and 6, the value before back-slash corresponds to the chunked files with 65\,000 entries(i.e., \texttt{MJD\_Train\_\{0-35\}.hdf5}); the value after back-slash corresponds to the last residual file (i.e., \texttt{MJD\_Train\_36.hdf5})}
\begin{tabular}{cccc}%{@{}l*{30}{l}}
\hline
\hline
Total No. of Data Points Size&&3\,193\,486&\\
\hline
Dataset Splitting&Train&Test&NPML\\
Splitting Ratio&75\%&20\%&5\%\\
\hline
File List[Full Set on DataPlanet]&\texttt{MJD\_Train\_\{0-36\}.hdf5}&\texttt{MJD\_Test\_\{0-9\}.hdf5}&\texttt{MJD\_NPML\_\{0-2\}.hdf5}\\
File List[Partial Set on Zenodo]&\texttt{MJD\_Train\_\{0-15\}.hdf5}&\texttt{MJD\_Test\_\{0-5\}.hdf5}&\texttt{MJD\_NPML\_\{0-2\}.hdf5}\\
No. of Data Points in Each File & 65\,000/55\,098& 65\,000/53\,691 & 65\,000/29\,697\\
HDF5 File Size&1.8Gb/1.6Gb&1.8Gb/1.6Gb&1.8Gb/862Mb\\
\hline
Dataverse Name:&&Physics&\\
Dataset Name:&&\textsc{Majorana Demonstrator} Experiment&\\
Persistent Identifier [DataPlanet]:&&perma:83.ucsddata/UQWQAV&\\
DOI[Zenodo]:&&\url{https://doi.org/10.5281/zenodo.8257027}&\\
\hline
\hline
\end{tabular}
\label{tab:ds_summary}
\end{table}
\end{center}

\par
\begin{figure}[htb!]
    \includegraphics[width=0.9\linewidth,,trim={0pc 0pc 0pc 0pc},clip]{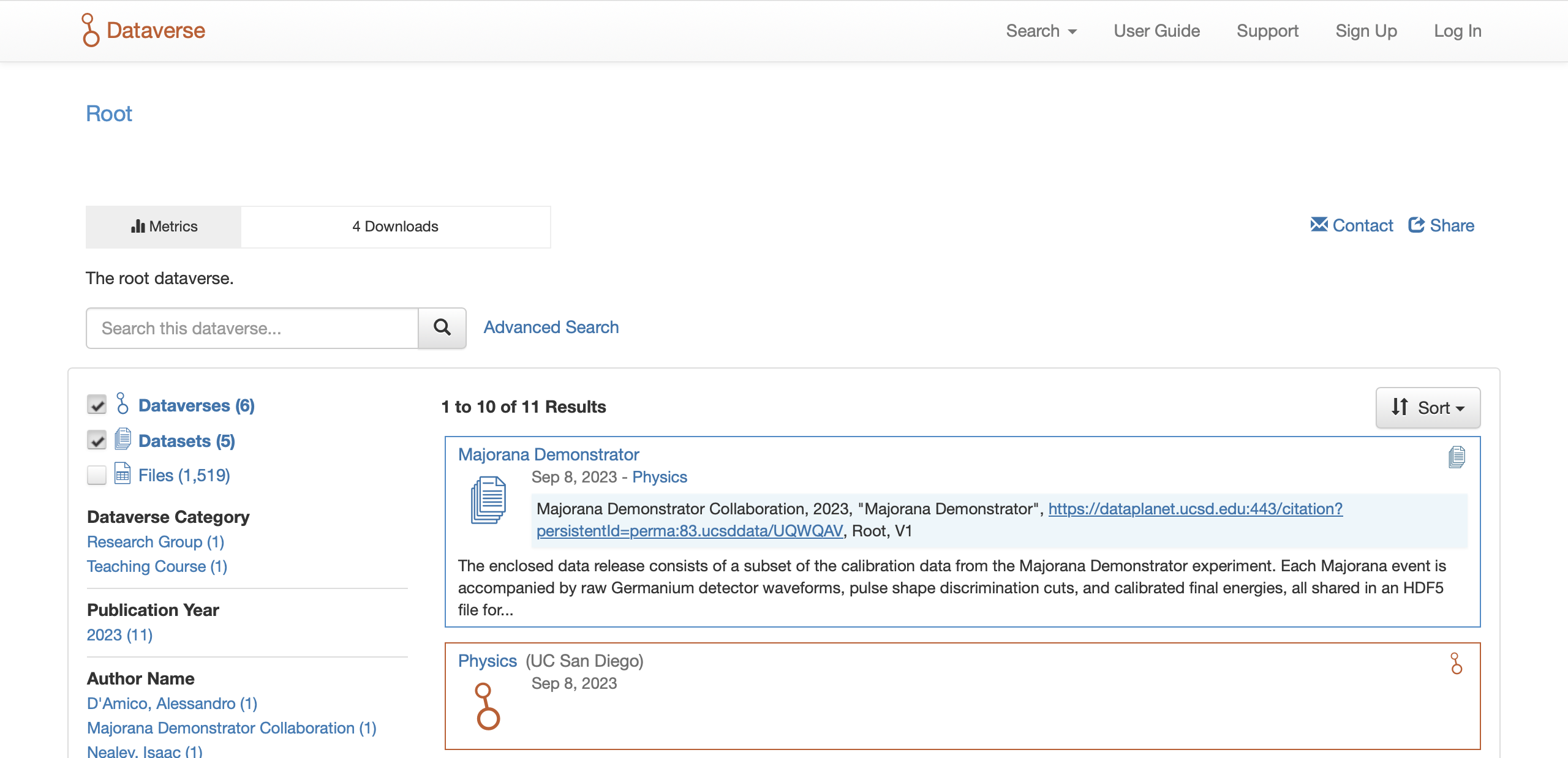}
    \caption{Left: the website screenshot of the DataPlanet interface, the ``Physics'' Dataverse and the ``\mjd~Experiment'' Dataset. The Dataset DOI is generated by the dataverse system.}
    \label{fig:phys_dataverse}
\end{figure}

 The open-sourced dataset will be stored at the San Diego Supercomputer Center (SDSC, \url{https://www.sdsc.edu}). The dataset will be made available using the DataPlanet software framework, an initiative by the UCSD Halıcıoğlu Data Science Institute (HDSI). To access the data, users should visit the DataPlanet website at \url{https://dataplanet.ucsd.edu}. In the list of dataverses, there is a dedicated Physics dataverse, as depicted in Figure~\ref{fig:phys_dataverse}. Alternatively, users can directly access the this dataset through the link: \url{https://dataplanet.ucsd.edu:443/citation?persistentId=perma:83.ucsddata/UQWQAV}. Under the Physics dataverse, users will find the ``Majorana Demonstrator Experiment'' dataset. After clicking into that dataset, users will see a comprehensive list of files detailed in Table~\ref{tab:ds_summary}.

%This platform enables public access to our data through web search, web download, and FTP transfer. Data Planet has a maximum file size limit of 2 Gb, prompting us to split the overall dataset into multiple hdf5 files smaller than 2\,Gb each to ensure compatibility. HDSI Data Planet comes with detailed user guide. The user should be able to follow the instruction and access these data as-need.
\par The DataPlanet system has recovered from its technical difficuties and is now fully operational. Besides, a partial release of the is made available on Zenodo. You can find the comprehensive list of files included in this pre-release in Table~\ref{tab:ds_summary}. This partial dataset is essentially the same as the complete DataPlanet dataset, with the exception that the train and test subsets contain fewer files. To access this partial dataset, please use the following \url{https://doi.org/10.5281/zenodo.8257027}. The complete dataset will become available once DataPlanet resumes its normal operations.

\section{NPML challenges}\label{sec:NPML}
\par The Neutrino Physics and Machine Learning (NPML) conference is scheduled to take place at Tufts University from August 22nd (Tuesday) to 25th (Friday), 2023. This event presents an excellent opportunity to showcase the dataset to the AI in physics community. On behalf of the organization committee, this document denotes the implementation detail of  "NPML Machine Learning Challenge" for the community. Participants in this challenge will be asked to design two machine learning models to accomplish two tasks:
\begin{enumerate}
    \item \textbf{Classification Task}: Develop a PSD ML Model that takes the  \texttt{raw\_waveform} as input and trains to accurately predict the 4 \texttt{psd\_label}s as denoted in Table~\ref{tab:osfield}. 
    \item Gather all events with a predicted \texttt{psd\_label==1}, i.e. event that simultaneously passes all the 4 PSD cuts. Call these events \textbf{clean events}.
    \item \textbf{Regression Task}: construct a Energy ML Model over all clean events. This model should take the corresponding \texttt{raw\_waveform} as input and be trained to accurately predict the associated \texttt{energy\_label}.
\end{enumerate}
\par
\begin{figure}[htb!]
    \centering
    \includegraphics[width=1.0\linewidth,,trim={8pc 1pc 8pc 1pc},clip]{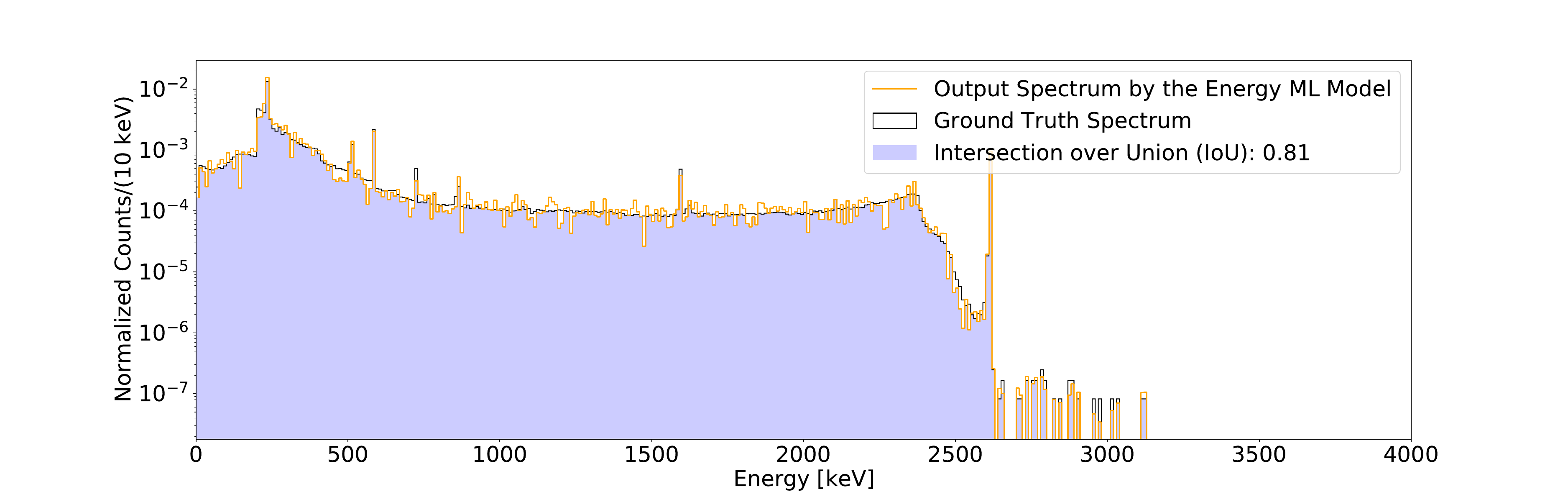}
    \caption{Illustration plot of ground truth spectrum, ML output spectrum and the calculated intersection over union.}
    \label{fig:iou_plot}
\end{figure}

 NPML challenge participants are requested to run both PSD ML model and Energy ML model over the challenge dataset, generate the following $(n,3)$ array, and send it to \url{liaobo77@ucsd.edu}:
\begin{center}
\renewcommand{\arraystretch}{1.7}
\setlength{\tabcolsep}{5pt}
\begin{table}[hbt]
\centering
\caption{An example of user-generated ML result over the NPML challenge dataset, need to be generated by participants.} 
\begin{tabular}{ccccc}%{@{}l*{30}{l}}
\hline
Dimension & & Content &  &Note\\
\hline
\texttt{id}& 0,1,2&\dots&159672,159673,159674&\\
predicted \texttt{psd\_label}& 0,1,0&\dots&1,0,1& 1 only if all 4 predicted \texttt{psd\_labels} are 1\\
predicted \texttt{energy\_label}& 3.25,1923.74,323.64&\dots&582.5, 938, 812.74& in keV\\
\hline
\end{tabular}
\label{tab:participant-array}
\end{table}
\end{center}
Upon receiving the array presented in Table~\ref{tab:participant-array}, the challenge organizer will conduct a comparison with the ground truth analysis label. This comparison will be quantified using two Intersection over Union (IoU) values, which serve as performance metrics for the model. IoU ranges from 0 (indicating complete non-overlapping) to 1 (indicating complete overlap). The first IoU will be computed between the ground truth \texttt{psd\_label} and the predicted \texttt{psd\_label}, providing a measure of the overlap between the two sets of labels. Next, clean events with \texttt{psd\_label==1} will be selected, and 1D histograms will be generated using \texttt{energy\_label}. This process is repeated for both the predicted and ground truth data. The second IoU will then be calculated by comparing the ground truth histogram with the predicted histogram, as shown in Figure~\ref{fig:iou_plot}. The final performance metric will be obtained by adding the two IoU values together.

\section{Disclaimer}\label{sec:disclamir}
The \textsc{Majorana Demonstrator} collaboration has authorized the public release of this dataset, permitting its use for all purposes. Individuals or collaborations are allowed to publish papers based on this dataset without including the \textsc{Majorana} collaborators as authors. The \textsc{Majorana Demonstrator} Collaboration maintains exclusive ownership rights over this dataset and reserves all associated rights. Should you choose to utilize this dataset in your work, the collaboration kindly requests that you properly cite both \cite{mjd_prl} and this ArXiv posting .
\begin{acknowledgments}
This material is based upon work supported by the U.S.~Department of Energy, Office of Science, Office of Nuclear Physics under contract / award numbers DE-AC02-05CH11231, DE-AC05-00OR22725, DE-AC05-76RL0130, DE-FG02-97ER41020, DE-FG02-97ER41033, DE-FG02-97ER41041, DE-SC0012612, DE-SC0014445, DE-SC0018060, DE-SC0022339, and LANLEM77/LANLEM78. We acknowledge support from the Particle Astrophysics Program and Nuclear Physics Program of the National Science Foundation through grant numbers MRI-0923142, PHY-1003399, PHY-1102292, PHY-1206314, PHY-1614611, PHY-1812409, PHY-1812356, PHY-2111140, and PHY-2209530. We gratefully acknowledge the support of the Laboratory Directed Research \& Development (LDRD) program at Lawrence Berkeley National Laboratory for this work. We gratefully acknowledge the support of the U.S.~Department of Energy through the Los Alamos National Laboratory LDRD Program, the Oak Ridge National Laboratory LDRD Program, and the Pacific Northwest National Laboratory LDRD Program for this work.  We gratefully acknowledge the support of the South Dakota Board of Regents Competitive Research Grant. 

We acknowledge the support of the Natural Sciences and Engineering Research Council of Canada, funding reference number SAPIN-2017-00023, and from the Canada Foundation for Innovation John R.~Evans Leaders Fund.  
We acknowledge support from the 2020/2021 L'Or\'eal-UNESCO for Women in Science Programme.
This research used resources provided by the Oak Ridge Leadership Computing Facility at Oak Ridge National Laboratory and by the National Energy Research Scientific Computing Center, a U.S.~Department of Energy Office of Science User Facility. We thank our hosts and colleagues at the Sanford Underground Research Facility for their support.
\end{acknowledgments}

\appendix
\section{Technical Details of Data Release}
\subsection{Dataset Detail}\label{app:detail}
\begin{itemize}
    \item Dataset: DS6 (1\% randomly subselected)
    \item Data Type: $^{228}$Th Calibration Run
    \item File Type: skim file
    \item Processing Tag: \texttt{GAT-v02-16-4-gd65d321}
    \item Run List:
    
25952 ; 27194 ; 28115 ; 28513 ; 28989 ; 31789 ; 32426 ; 32869 ; 34277 ; 34501 ; 34555 ; 34591 ; 35561 ; 35586 ; 35649 ; 35944 ; 36929 ; 37070 ; 38017 ; 39220 ; 39241 ; 39342 ; 41071 ; 41119 ; 41171 ; 41571 ; 42491 ; 42492 ; 42876 ; 43662 ; 43978 ; 43982 ; 43983 ; 48944 ; 50436 ; 50659 ; 50928 ; 51289 ; 51470 ; 51664 ; 52510 ; 52673 ; 52678 ; 52713 ; 54328 ; 55749 ; 55891 ; 56056 ; 56301 ; 57365
    \item Total Number of Events: 3\,193\,486
    \item Input files: can be composed from the above information using 
    \begin{verbatim}
        os.path.join("/global/cfs/cdirs/majorana/data/mjd/surfmjd/analysis/skim/", 
        "DS6cal", "GAT-v02-16-4-gd65d321", 
        "skimDS6_run{run_umber}_small_s.root")
    \end{verbatim}
\end{itemize}
\subsection{MJD Data Extraction}
The data processing code loops through all the runs listed above. For each run, we use uproot~\cite{uproot} to read information from the skim file and store those as an awkward array~\cite{awkarray}.  we first apply the suite of data cleaning and run selection (DSRC) cut listed below:
\begin{verbatim}
    np.all(ev["isGood"]) 
    and (not (ev["isLNFill1"]&ev["C"][chindex]==1)) 
    and (not (ev["isLNFill2"]&ev["C"][chindex]==1)) 
    and np.all(ev["wfDCBits"])==0 and ev["muVeto"]==0 
    and (ev["mH"]==1 and ev["mL"]==1)
\end{verbatim}
Although not specified, the production of the skim file uses a sumE$>$100\,keV cut, which results in a step-like feature around 100\,keV in Figure~\ref{fig:spectrum}. For each event passing all DSRC cuts, we extract information listed in Table~\ref{tab:osfield} in the following way:
\begin{itemize}
    \item \texttt{raw\_waveform} is not directly read out from the root file, but rather with the following code:
    \begin{verbatim}
        ds = GATDataSet(int(run))
        tt_gat = gds.GetGatifiedChain()
        tt_blt = gds.GetBuiltChain()
        tt_gat.AddFriend(tt_blt)
        tt_gat.GetEntry(0)
        is_ms = tt_blt.run.GetUseMultisampling()
        if is_ms:
            wfdown = tt_gat.event.GetWaveform(iH) # downsampled
            wffull = tt_gat.event.GetAuxWaveform(iH) # fully sampled
            wf = MJTMSWaveform(wfdown, wffull)
        else:
            wf = tt_gat.event.GetWaveform(iH)
    \end{verbatim}
    As demonstrated within the code, the waveform readout is accomplished with the Majorana/GERDA Data Objects (MGDO) Library. Initially scripted in C++, MGDO can be conveniently accessed through the pyROOT interface of the customized \textsc{Majorana} ROOT. The \texttt{run} number comes from the looping. \texttt{iE}, \texttt{iH} are obtained from the skim file of the corresponding \texttt{run}. \texttt{is\_ms} is true if the current run uses multisampling and false otherwise. If multisampling is used, the waveforms generated by \texttt{MJTMSWaveform(wfdown, wffull)} will contain fully-sampled rising edge and linearly-interpolated baseline region and decay tail. The linear interpolation is implemented every 4 samples. The extracted waveforms can be found in Figure~\ref{fig:mjdwf}.
    \item \texttt{energy\_label} is directly extracted from the \texttt{Final\_Energy} field in skim file.
    \item  \texttt{psd\_label\_low\_avse} is generated based on \texttt{avse\_corr} field in the skim file. The label reads 1 if \texttt{avse\_corr} is greater than -1.0 and 0 otherwise.
    \item  \texttt{psd\_label\_high\_avse} is generated based on \texttt{avse\_corr} field in the skim files, \texttt{corr} means drift time corrected. The label reads 1 if \texttt{avse\_corr} is smaller than 9.0 and 0 otherwise.
    \item  \texttt{psd\_label\_dcr} is generated based on \texttt{dcr\_corr} field in the skim file. The label reads 1 if \texttt{dcr\_corr} is smaller than 2.326 and 0 otherwise.
    \item  \texttt{psd\_label\_lq} is generated based on \texttt{lq} field in the skim file. The label reads 1 if \texttt{lq} is smaller than 5.0 and 0 otherwise.
    \item \texttt{tp0} is obtained from this line of code:\begin{verbatim}
        correctedT0 = tt_gat.correctedT0.at(iH)
        tp0 = int(np.round((correctedT0 + 2140)/10)) 
    \end{verbatim}. Note that \texttt{tt\_gat} is created when obtaining \texttt{raw\_waveform}. The second line is used to converted correctedT0~(in unit of ns) to the actual index of the waveform. 2140 corresponds to the width of trapezoidal filter.
    \item  \texttt{detector} is the unique detector id. It is generated by combining the CPD value with string concatenation. For example, if a detector is located in C2P7D2, its corresponding \texttt{detector} id will reads 272.
    \item  \texttt{run\_number} is the original run number for each file, as specified in Section~\ref{app:detail}.
    \item  \texttt{id} is the unique id per waveform/datapoint. It is generated when we create the HDF5 file, as we will discuss in Section~\ref{app:hdf5gen}
\end{itemize}
\subsection{HDF5 File Generation}\label{app:hdf5gen}
After obtaining all data points and its associated fields listed above, the next step is to sort them into train, test, and NPML dataset. The detailed procedure is described below:
\begin{enumerate}
    \item Loop through each run
    \item For each run, randomly split its data points into 3 subsets with 75\%:20\%:5\% ratio, and append them into train, test and NPML dataset, respectively.
    \item finish looping through all the runs
    \item Separately shuffle the events within train, test, and NPML dataset
    \item Assign a unique integer id to each data point within the 3 datasets. The id of the train, test, and NPML dataset ranges from [0,2\,395\,098), [2\,395\,099,3\,033\,789), and [3\,033\,790,3\,193\,486) respectively.
    \item Chunk the train dataset into 37 hdf5 files with name \texttt{MJD\_Train\_\{0-36\}.hdf5}, each containing 65\,000 data points. \texttt{MJD\_Train\_36.hdf5} will contain residual events thereby less than 65\,000 events.
    \item  Repeat step 5 on the test dataset to get \texttt{MJD\_Test\_\{0-9\}.hdf5}
    \item  Repeat step 5 on the NPML dataset to get \texttt{MJD\_NPML\_\{0-3\}.hdf5}
\end{enumerate}
Step 2 guarantees that train, test, and NPML dataset will all contains events from the entire run range; Step 4 guarantees that each chunk will contain events from the entire run range. The chunk size of 65\,000 is selected to constrain the size of a single \texttt{hdf5} file: the UCSD DataPlanet system limit file size to 2\,Gb, and a \texttt{hdf5} file with 65\,000 data points is roughly 1.8\,Gb. \texttt{MJD\_Train\_36.hdf5}, \texttt{MJD\_Test\_9.hdf5}, and \texttt{MJD\_NPML\_2.hdf5} will contain the residual amount of events within each dataset, thereby the amount of data points is smaller than 65\,000 and the file size is also smaller. \texttt{MJD\_Train\_36.hdf5} contains 55\,098 events corresponding to 1.6Gb; \texttt{MJD\_Test\_9.hdf5} contains 53\,691 events corresponding to 1.6Gb; \texttt{MJD\_NPML\_2.hdf5} contains 29\,697 events corresponding to 862Mb.
\nocite{*}

\bibliography{BDT_paper}% Produces the bibliography via BibTeX.

\end{document}